\documentclass[10pt,twocolumn,letterpaper]{article}

\usepackage{wacv}              

\usepackage{graphicx}
\usepackage{amsmath}
\usepackage{amssymb}
\usepackage{booktabs}

\usepackage[pagebackref,breaklinks,colorlinks]{hyperref}

\usepackage{orcidlink}
\usepackage{times}
\usepackage{helvet}
\usepackage{courier}
\usepackage{amsfonts}
\usepackage{algorithmic}
\usepackage{algorithm}
\usepackage{array}
\usepackage{textcomp}
\usepackage{stfloats}
\usepackage{url}
\usepackage{verbatim}
\usepackage{epsfig}
\usepackage{makecell}
\usepackage{multirow}
\usepackage{pifont}
\usepackage{tablefootnote}
\usepackage{enumitem}
\usepackage{wrapfig}

\usepackage{amsmath}

\usepackage{caption}
\usepackage{subcaption}
\usepackage{tikz}
\usetikzlibrary{shapes,decorations,arrows,calc,arrows.meta,fit,positioning}

\usepackage[capitalize]{cleveref}
\crefname{section}{Sec.}{Secs.}
\Crefname{section}{Section}{Sections}
\Crefname{table}{Table}{Tables}
\crefname{table}{Tab.}{Tabs.}

\def\wacvPaperID{2339} 
\def\confName{WACV}
\def\confYear{2025}

\begin{document}

\title{Who Brings the Frisbee: Probing Hidden Hallucination Factors in Large Vision-Language Model via Causality Analysis}

\author{Po-Hsuan Huang$^{1*}$, Jeng-Lin Li$^{1*}$, Chin-Po Chen$^1$, Ming-Ching Chang$^{1, 2}$, Wei-Chao Chen$^1$\\
$^1$Inventec Corporation, $^2$ University at Albany - SUNY\\
$^1$No. 66, Hougang St., Shihlin Dist., Taipei City, $^2$ Albany, NewYork\\
{\tt\small $^1$\{huang.reese, li.johncl, chen.jackcp, chang.ming-ching, chen.wei-chao\}@inventec.com}, \\ {\tt\small $^2$mchang2@albany.edu}
\thanks{The first two authors equally contribute to this work.}
}
\maketitle

\begin{abstract}
Recent advancements in large vision-language models (LVLM) have significantly enhanced their ability to comprehend visual inputs alongside natural language. However, a major challenge in their real-world application is hallucination, where LVLMs generate non-existent visual elements, eroding user trust. The underlying mechanism driving this multimodal hallucination is poorly understood. Minimal research has illuminated whether contexts such as sky, tree, or grass field involve the LVLM in hallucinating a frisbee.
We hypothesize that hidden factors, such as objects, contexts, and semantic foreground-background structures, induce hallucination. This study proposes a novel causal approach: a hallucination probing system to identify these hidden factors. By analyzing the causality between images, text prompts, and network saliency, we systematically explore interventions to block these factors. Our experimental findings show that a straightforward technique based on our analysis can significantly reduce hallucinations. Additionally, our analyses indicate the potential to edit network internals to minimize hallucinated outputs.
\end{abstract}


\section{Introduction}
\label{sec:intro}

Large vision-language models (LVLM) can comprehend multimodal data and respond to human commands~\cite{dai2024instructblip,ye2023mplug,zhu2023minigpt,liu2023llava}. Alongside advancements in network architectures, significant research focuses on improving response accuracy and reducing deviations from human instructions~\cite{dai2024instructblip,liu2023llava}. 
Despite these efforts, modern LVLMs struggle with real-world challenges due to their notorious hallucinations~\cite{liu2024survey,bai2024hallucination} jeopardizing downstream reliability and safety.

LVLM hallucinations occur when the generated contents do not align with the provided visual cues or include unrelated or incorrect texts~\cite{liu2024survey}. Mitigating hallucinations by fine-tuning LVLMs with human preferences is effective but expensive, requiring extensive human annotations~\cite{yu2023rlhf}. To reduce costs, recent research employs auxiliary models, such as object grounding and language refinement models, to automatically generate pseudo annotations~\cite{yu2023hallucidoctor}.
Alternatively, approaches that require LVLMs to answer multiple verification questions iteratively incur significant computational overhead~\cite{wu2024logical}. Besides directly optimizing human preferences, post-training inference calibration, for example, enforcing the decoding with contrast to erroneous variations, can partly reduce hallucination probability~\cite{leng2023mitigating}.

\begin{figure*}
\centerline{
  \includegraphics[width=\linewidth,height=0.25\linewidth]{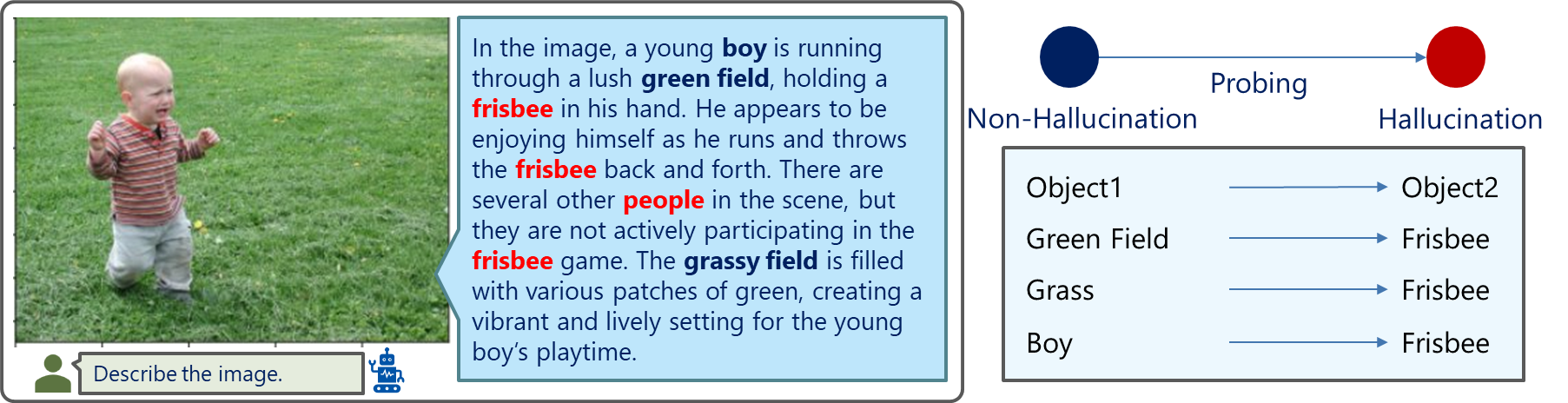} 
  \vspace{-2mm}
}
\caption{The InstructBLIP~\cite{dai2024instructblip} LVLM hallucinates a frisbee when describing a boy in the green field. There might be a spurious correlation between a boy and a frisbee. Meanwhile, the green field is another non-hallucinatory subject that might induce hallucinations. This underlying relation remains underexplored in the hallucination reduction research.}
\label{fig:teaser}
\vspace{-3mm}
\end{figure*}

Despite various strategies proposed to reduce hallucination in LVLMs, a limited understanding of their response behaviors still hinders further research. Clearly, various uncontrolled hidden factors contribute to these intricate hallucinations when the LVLMs process multi-modality data. For example, Figure~\ref{fig:teaser} shows an example where InstructBLIP~\cite{dai2024instructblip} erroneously describes a nonexistent frisbee. This mistake likely arises from the presence of a large green grass field in the photo, where frisbees frequently co-occur in the training data. Spurious correlations lead to cross-modality retrieval errors, where models predict objects based on their frequent occurrences in training data~\cite{Kim_2023_CVPR}. Mitigating these errors requires fine-grained data augmentation and balancing across modalities~\cite{biten2022let}. Additionally, poor image-text alignment biases the decoding mechanism towards language, which tends to neglect image contents~\cite{jiang2023hallucination,leng2023mitigating}. Analytical studies, such as~\cite{zhou2024analyzing}, have identified factors like occurrence, uncertainty, and object position through statistical analyses, but lack a unified framework to analyze the hidden factors inducing hallucination. These studies have not scrutinized hidden context factors, such as the people, trees, or grass in the image, that might induce hallucinations.

\vspace{-2mm}

It is crucial to understand the conditions that lead to LVLM hallucinations and analyze their {\bf causal patterns}. When there are green fields, it is likely to have a frisbee in the training data, leading to the hallucination mentioned above in Figure~\ref{fig:teaser}. 
However, the reverse scenario is less likely, due to the relatively fewer training data focusing mainly on frisbees. Additionally, there is a significant gap in studies connecting hallucination analysis with hallucination reduction strategies. To address this, we systematically investigate four fundamental research questions about LVLM hallucination concerning visual objects: (1) Are semantic structures affecting hallucination? (2) What are the effects of non-hallucinated objects that are potentially accompanied by hallucinated objects? (3) How likely can we intervene in LVLM regarding hallucinated objects to reduce the effects of hidden factors? (4) Are there salient network internals implying network hallucination?



This study explores object hallucination patterns and introduces a novel causal intervention scheme to analyze LVLM behaviors. 
We examine the causal relations between visual objects in the dataset and observe the generated LVLM outputs, overviewed in Figure~\ref{fig:editing}. 
Our findings highlight the key relations between a main subject and the context. 
We demonstrate that simple intervention to the observed structures can notably reduce hallucinations. 
Additionally, we investigate the network internal saliency of latent embeddings based on causally-related inputs. Our analysis results pave the way for seamless interference in model generation to mitigate hallucination occurrences.

Our contributions are summarized in the following:
\vspace{-2mm}
\begin{itemize}[leftmargin=10pt] \itemsep -.15em
\item Investigating hallucination relations between hallucination-inducing words and hallucinatory words 
\item Identifying a unified causality graph to develop hallucination reduction solutions 
\item Analyzing intervention on text, image, and embedding
\item Probing embedding properties of high-hallucinatory and non-hallucinatory images
\end{itemize}

\vspace{-2mm}
\section{Related Work}
\vspace{-1mm}

\vspace{-1mm}
\subsection{Large Vision-Language Model (LVLM)}
\vspace{-1mm}

Large pre-trained models have heralded a new era of vision language models. InstructBLIP~\cite{dai2024instructblip}, mPLUG-Owl~\cite{ye2023mplug1,ye2023mplug}, MiniGPT~\cite{zhu2023minigpt}, and LLaVa~\cite{liu2023llava} all leverage autoregressive pre-trained large language models (LLMs) paired with fine-tuned vision encoders and multimodal alignment modules. Researchers construct extensive instruction datasets for instruction tuning, aiming to enhance response quality~\cite{liu2023llava}. Additionally, automatic instruction generation approaches can effectively scale up training~\cite{wang2024vigc}. Despite addressing known shortcomings of models, challenges persist in fully understanding their behaviors.

\vspace{-1mm}
\subsection{Hallucination Detection and Elimination}
\vspace{-1mm}

Mitigating hallucination issues was initially studied in the field of image captioning and later extended to LVLMs. Traditional image captioning metrics such as CIDEr~\cite{vedantam2015cider} and METOER~\cite{banerjee2005meteor} often fail to capture the object hallucination within a sentence. 
The polling-based hallucination detection method (POPE) involves a large set of questions asking whether a specific object is present in an image~\cite{li2023evaluating}.  
Despite the LVLM evaluation rapidly expanding to various types of hallucinations~\cite{xu2023lvlm,qiu2024valor,sun2023aligning}, the fundamental issue of visual object hallucination remains unsolved.

\textbf{Instruction tuning:} 
LRV-Instruction~\cite{liu2023mitigating} dataset is designed to balance positive and negative instructions for robust instruction tuning. Direct preference optimization improves the model using annotated data of both hallucinated and non-hallucinated samples, supported by a reward model providing feedback during fine-tuning~\cite{gunjal2024detecting}. However, the quality of collected instructions and the computational overhead continue to be major inconveniences.

\textbf{Self-check and auxiliary models:}
Volcano~\cite{lee2023volcano} is a self-feedback guided revision model that iteratively asks itself questions to improve response quality. To mitigate the self-bias of LVLMs leading to hallucinations, object existence verification can be achieved by prompting another open-vocabulary object detector~\cite{yan2024vigor,zhao2024mitigating}. These verification results serve as a reward model during LVLM fine-tuning. 
More complex multiple-step strategies including question generation, object grounding, and language refinement are subsequently designed~\cite{yin2023woodpecker,yu2023hallucidoctor}.
These methods avoid updating model parameters, relying on regenerated or refined answers from ChatGPT~\cite{achiam2023gpt}, indicating their success hinges on the strong summarization ability of ChatGPT. We observe that discriminative ability in vision question answering does not directly indicate generative hallucination (reported in supplementary). Hence, we focus on hallucination in the generative task for our framework design.

\textbf{Post-training adjustment methods:}
Recent studies investigate various decoding strategies, including {\em contrastive decoding}~\cite{leng2023mitigating} and beam search strategies~\cite{huang2023opera}. VCD~\cite{leng2023mitigating} calibrates output predictions by contrasting the output distribution between original and distorted visual inputs.
OPERA~\cite{huang2023opera} explores related patterns in attention outputs and removes candidate words exhibiting aggregation patterns to prevent decoding hallucination words.
Some previous works investigate post-training {\em calibration} methods to avoid the cost of fine-tuning. For instance, Lin et al.~\cite{lin2024revisiting} calibrates predictions by estimating linguistic bias in the training data. While recent LLM studies address hallucinations through probing and editing strategies~\cite{li2023inference,chen2024truth}, these approaches have not been explored in LVLMs.

\textbf{Causality} provides a robust framework for uncovering hidden inference structures. 
The Structural Causal Model (SCM)~\cite{pearl2009causality} has been widely employed in debiasing LLMs to unveil inherent causal relations between data and labels~\cite{zhou2023causal}.
A causal attention mechanism was developed to mitigate unobserved confounding factors in vision-language Transformers~\cite{yang2021causal}. Additionally, causal effects are also integrated into loss functions to address counterfactual biases~\cite{vosoughi2024cross}. However, the hallucination phenomenon in LVLM remains under-investigated due to the naturally complex multimodal structure, hindering systematic understanding. Most previous studies focus on vision-language discriminative tasks with clear binary targets~\cite{palit2023towards,chen2023deconfounded,lovenia2023negative} or short generated sequences in image captioning tasks~\cite{yang2021deconfounded,liu2022show}. There is a lack of exploration into causal structures for hallucination patterns in generative tasks.



\vspace{-2mm}
\section{Method}
\vspace{-1mm}

\tikzset{
    -{Latex[length=1.45mm]},auto,node distance = 0.6 cm and 0.6 cm,semithick,
    state/.style ={ellipse, draw, minimum size=0.3cm},
    point/.style = {circle, draw, inner sep=0.01cm,fill,node contents={}},
    bidirected/.style={Latex-Latex,dashed},
    el/.style = {inner sep=1pt, align=left, sloped},
    cross/.style={
            draw=none,
            path picture={
                \draw[red, thick, -]
                (path picture bounding box.south east) -- (path picture bounding box.north west)
                (path picture bounding box.south west) -- (path picture bounding box.north east);
            }
        }
}

\begin{figure*}[t]
\centerline{
     \begin{subfigure}[b]{0.16\textwidth}
         \centering
         \begin{tikzpicture}[scale=0.8, every node/.style={transform shape}]
            \node[state] (i) at (0,0) { $X$};
            \node[state] (zo) [right = of i] {$Z_o$};
            \node[state] (a) [right = of zo] {$A$};
            \path (i) edge (zo);
            \path (zo) edge (a);
        \end{tikzpicture}
         \caption{}
         \label{fig:causality_graphs_ideal}
     \end{subfigure}
     \hspace{0.3cm} 
     \begin{subfigure}[b]{0.16\textwidth}
         \centering
        \begin{tikzpicture}[scale=0.8, every node/.style={transform shape}]
            \node[state] (i) { $X$};
            \node[state] (zo) [right =of i] { $Z_o$};
            \node[state] (zc) [below =of zo] { $Z_c$};
            \node[state] (a) [right =of zo] { $A$};
            \path (i) edge (zo);
            \path (i) edge (zc);
            \path (zo) edge (a);
            \path (zc) edge (a);
            \path (zc) edge (zo);
        \end{tikzpicture}
         \caption{}
         \label{fig:causality_graphs_general}
     \end{subfigure}
    \hspace{0.3cm} 
    \begin{subfigure}[b]{0.16\textwidth}
         \centering
        \begin{tikzpicture}[scale=0.8, every node/.style={transform shape}]
            \node[state] (i) { $I$};
            \node[state] (zo) [right =of i] { $Z_o$};
            \node[state] (zc) [below =of zo] { $Z_c$};
            \node[state] (a) [right =of zo] { $A$};
            \path (i) edge (zo);
            \draw[red, thick, - ] ($(zo)!0.5!(zc)$) ++(-0.1, -0.1) -- ++(0.2, 0.2);
            \draw[red, thick, - ] ($(zo)!0.5!(zc)$) ++(-0.1, 0.1) -- ++(0.2, -0.2);
            \path (zo) edge (a);
            \path (zc) edge (a);
            \path (zc) edge (zo);
        \end{tikzpicture}
         \caption{}
         \label{fig:causality_graphs_image}
    \end{subfigure}
    \hspace{0.3cm} 
    \begin{subfigure}[b]{0.20\textwidth}
         \centering
        \begin{tikzpicture}[scale=0.8, every node/.style={transform shape}]
            \node[state] (i) { $Q$};
            \node[state] (zo) [right =of i] { $Z_o$};
            \node[state] (zc) [below =of zo] { $Z_c$};
            \node[state] (s) [right = of zo] { $S$};
            \node[state] (a) [right =of s] { $A$};
            \path (i) edge (zo);
            \path (zo) edge (s);
            \path (zc) edge (a);
            \path (zc) edge (zo);
            \path (s) edge (a);
            \draw[red, thick, - ] ($(zo)!0.5!(zc)$) ++(-0.1, -0.1) -- ++(0.2, 0.2);
            \draw[red, thick, - ] ($(zo)!0.5!(zc)$) ++(-0.1, 0.1) -- ++(0.2, -0.2);
        \end{tikzpicture}
         \caption{}
         \label{fig:causality_graphs_text}
    \end{subfigure}
    \hspace{0.3cm} 
    \begin{subfigure}[b]{0.14\textwidth}
         \centering
        \begin{tikzpicture}[scale=0.8, every node/.style={transform shape}]
            \node[state] (zo) { $Z_o$};
            \node[state] (zc) [below =of zo] { $Z_c$};
            \node[state] (a) [right =of zo] { $A$};
            \path (zc) edge (a);
            \path (zc) edge (zo);
            \path (zo) edge (a);
            \draw[red, thick, - ] ($(zo)!0.5!(zc)$) ++(-0.1, -0.1) -- ++(0.2, 0.2);
            \draw[red, thick, - ] ($(zo)!0.5!(zc)$) ++(-0.1, 0.1) -- ++(0.2, -0.2);
        \end{tikzpicture}
         \caption{}
         \label{fig:causality_graphs_embedding}
    \end{subfigure}
\vspace{-2mm}
}    
\caption{
(a) Ideal LVLM generation. 
(b) The {\bf causal graphical model for LVLM generation}. 
(c-e) Deconfounded by (c) image intervention, (d) text intervention, and (e) embedding intervention.}
\label{fig:causality_graphs}
\vspace{-4mm}
\end{figure*}
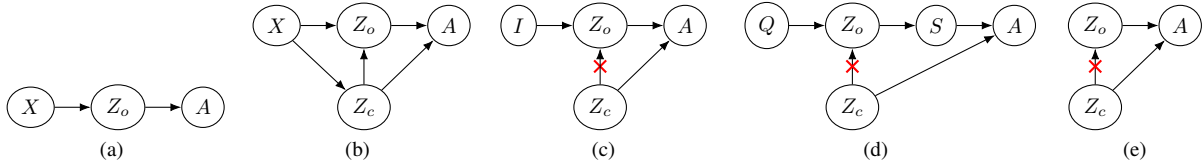


\subsection{Datasets, LVLM Setup, and Hallucination Evaluation Metrics}
\label{ssec:setup}

{\bf Datasets:}
We utilize the AMBER dataset~\cite{wang2023llm} and COCO dataset~\cite{lin2014microsoft} for evaluation. AMBER is a benchmark dataset assessing LVLM hallucination, including human annotations on both truly appeared objects and potentially hallucinated ones.
The setting of COCO is followed by prior works~\cite{huang2023opera} with randomly selected 500 images from the 2014 validation set. Here, we prompt LVLMs with ``Describe this image.'' to obtain descriptions for the images.


\textbf{Large vision-language model (LVLM):}
We utilize an auto-regressive transformer-based LVLM denoted as $f$. The input token sequence $X\in \mathbb{R}^{N\times T}$ with $N$ samples and $T$ timesteps is fed into the LVLM to the latent embedding $E\in \mathbb{R}^{N\times T\times D}$ for $D$ latent dimensions and generate the text $A=f(X)$. We employed two established LVLMs including InstructBLIP~\cite{dai2024instructblip} and mPLUG-Owl2~\cite{ye2023mplug}.

\textbf{Hallucination evaluation metrics:}
The evaluation function $H$ assigns a hallucinatory score $H(A)\in \mathbb{R}$ to a response $A$, indicating hallucination rates with given metrics. We follow the metrics used in the dataset papers for evaluation. 
We assess LVLM hallucination using several common metrics: CHAIR, Cover, HAL, and Cog scores on AMBER dataset~\cite{wang2023llm} as well as \text{$CH_i$}, \textit{$CH_s$}, and Recall on COCO~\cite{huang2023opera}.
CHAIR~\cite{chair2018} measures the hallucination generation rate. \textit{$CH_i$} and \textit{$CH_s$}, represent the image-level and sentence-level hallucination rates, respectively. Cover and Recall measure the ratio of mentioned existing objects to all existing objects in the image. HAL indicates whether the CHAIR score is non-zero for a sentence. Cog measures the hallucinatory object rate resembling human cognition.
\begin{table*}[t]
\centerline{
\normalsize{
\begin{tabular}{c|c|c|c|c|c||c|c|c|c|c|c}
\toprule
\multicolumn{6}{c||}{InstructBLIP}  & \multicolumn{6}{c}{mPLUG-Owl2} \\
\midrule
(a) $O_h$  & \# & (b) $O_{h}^{1}, O_{h}^{2}$ & \# & (c) $O_{n}$ & \# & (a) $O_h$  & \# & (b) $O_{h}^{1}, O_{h}^{2}$ & \# & (c) $O_{n}$ & \# \\ 
\midrule
                              people & 68    & (people, $O_{h}^{2}$)  & 68    & tree   & 60   & people & 88    & (people, $O_{h}^{2}$)  & 40    & water  & 41    \\
                              person & 34    & (person, $O_{h}^{2}$)  & 41    & water  & 52    & bottle & 40    & (cup, $O_{h}^{2}$)     & 40    & tree   & 39    \\ 
                              car    & 25    & (cup, $O_{h}^{2}$)     & 26    & sky    & 32   & car    & 38    & (bottle, $O_{h}^{2}$)  & 39    & road   & 38    \\ 
                              tree   & 21    & (bottle, $O_{h}^{2}$)  & 25    & beach  & 31    & person & 31    & (person, $O_{h}^{2}$)  & 39    & beach  & 32    \\
                              sun    & 21    & (bicycle, $O_{h}^{2}$) & 22    & road   & 31   & chair  & 24    & (book, $O_{h}^{2}$)    & 30    & people & 27    \\
\bottomrule

\end{tabular}%
}
\vspace{-2mm}
}
\caption{This table shows the counts of the top 5 frequent (a) {\em single-hallucinatory} words $O_h$, (b) {\em co-occurring hallucinatory} words ($O_{h}^{1}, O_{h}^{2}$) (c) {\em Hallucinatory-inducing} words $O_n$. $O_n$ is frequently associated with other hallucinatory words $O_h$, i.e., $O_n \rightarrow O_{h}$. \# denotes the counts of the word. See Supplementary for full results.}
\label{tab:Cooccurance_count}
\vspace{-4mm}
\end{table*}
\subsection{Hallucination Statistics}
\label{ssec:h_statistics}
We investigate the hallucination results using human-annotated labels from the AMBER dataset to illustrate the underlying data and model properties. Table~\ref{tab:Cooccurance_count} lists the top 5 most common hallucination cases, including a {\em single-hallucinatory} word, {\em co-occurring hallucinatory} words, and the {\em hallucinatory-inducing} words. The hallucinatory-inducing words are non-hallucinatory while associated with other hallucinatory words. These cases were detected using InstructBLIP and mPLUG-Owl2 on the AMBER dataset, suggesting that LVLMs exhibit different hallucination inclinations.



\textbf{Co-occuring hallucinatory words:}
Frequently hallucinated words tend to co-occur with other hallucinated words. Once a hallucination occurs, other hallucinations are likely to follow. We present the top 5 most common co-occurring hallucination words in Table~\ref{tab:Cooccurance_count} (b). Observe that co-occurring hallucinatory words differ from single-hallucinatory words. The appearance of these paired hallucinated words is usually syntactically correct in image descriptions, which might be one reason that paired cooccurrence words appear. These observations suggest that the underlying syntactic and semantic structures influence the relationships between words of hallucinations.




\textbf{Hallucinatory inducing words:}
Aside from the co-occurrence of hallucinatory words, we are also interested in an important causal factor ``What induces a hallucination?''. Given a generated response $A$, the mentioned object set is denoted as $S = \{s_1,\ldots,s_n\}$ and the ground-truth object set is $O = \{o_1,\ldots,o_m\}$. 
The hallucinatory words are the objects absent in the ground-truth object set, i.e., $O_{h} = S \setminus O$, and the set of non-hallucinatory words is denoted as $O_n=S \cap O$. We regard the unrevealed relation between hallucinatory and non-hallucinatory words using conditional probability $P(O_{h} \mid O_n)$~\cite{Pearl2019TheST}. Table~\ref{tab:Cooccurance_count} (c) shows the $O_n \rightarrow O_{h}$ relations in AMBER. Words like `tree', `water', `sky', `beach', and `road' are scene-like nouns, which usually describe the background information. The appearance of these types of nouns often induces other hallucinatory words. For example, a tree induces a bicycle to hallucinate probably because they are likely to appear in the same scene.
However, the observed hallucination word relations remain unclear regarding causality, raising concerns about spurious correlations and risks of misinterpretation. Therefore, we utilize the causality analysis to identify solutions for hallucination reduction.
\vspace{-2mm}
\begin{figure*}
\centerline{
  \includegraphics[width=0.95\linewidth,height=0.35\linewidth]{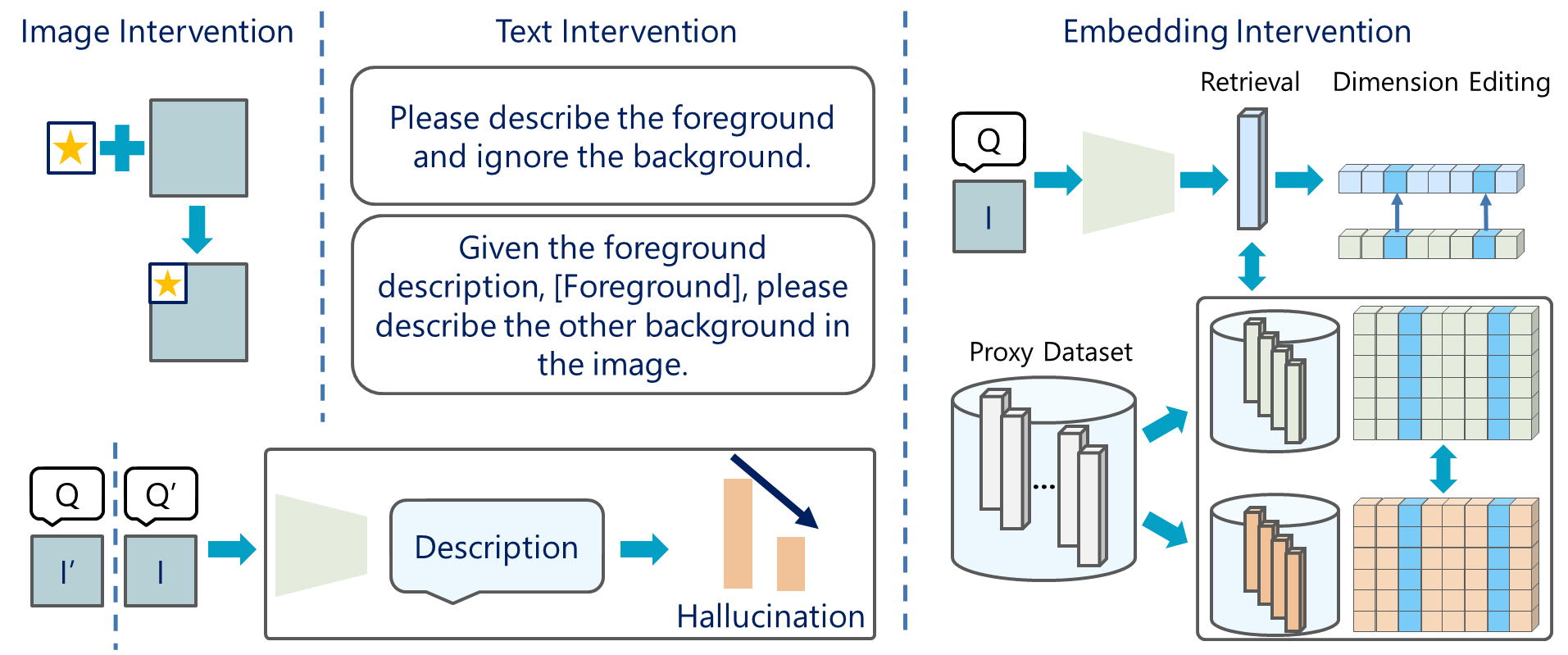} 
  \vspace{-2mm}
}
\caption{Our proposed image, text, and embedding intervention approaches correspond to Figure~\ref{fig:causality_graphs} (c), (d), and (e).}
\label{fig:editing}
\vspace{-2mm}
\end{figure*}

\subsection{Causality Analysis}
\label{ssec:method_causality_analysis}

We formulate a causal graphical model involving the input random variables: image $I$, text query $Q$, a latent variable of target object $Z_o$, context factor $Z_c$, and the resulting answer $A$. This model is represented by a directed acyclic graph (DAG) as shown in Figure~\ref{fig:causality_graphs}. 
In this graph, a directed edge between variables indicates a direct causal influence of the parent node on the child node.

We distinguish and abstract the variables $Z_o$ and $Z_c$ at a cognitive level. $Z_o$ represents the ideal semantic representation of target objects ({\em e.g.}, the concept of a car), while $Z_c$ as a {\bf confounding} variable denotes a context pattern that could diversify the comprehension of the car. 
Figure~\ref{fig:causality_graphs_ideal} depicts an ideal LVLM generation, where $A$ is independent of $Z_c$. However, the inherent bias in the training data introduces $Z_c$ into the {\bf causal graphical model for LVLM generation} in Figure~\ref{fig:causality_graphs_general}, shaping the unwanted causal effect $Z_c\rightarrow Z_o$.




%


We adopt the standard causality analysis approach~\cite{pearl2016causal,pearl2009causality}; performing an intervention on input nodes to block the undesired effects from the confounding factor $Z_c$.
We set up the causal effect metric $\delta(P, P')$ between a distribution $P$ and the distribution after intervention $P'$ is defined as the fewer hallucinations after the intervention; that is, $\delta = \mathbb{I}(H(A) > H(A'))$, where $A'$ denotes the resulting answer after the intervention and $H$ is the hallucination evaluation metric defined in $\S$\ref{ssec:setup}. The causal effect metric $\delta$ reflects whether an intervention successfully reduces hallucination. We measure the {\bf total causal effect (TCE)} ~\cite{pearl2001direct,stolfo2023causal} over an evaluation test set. For the intervention results, TCE calculates the expected value $\mathbb{E}$ of $\delta$ over the evaluation test set $X$ by the equation:
\begin{equation}
    TCE = \mathbb{E}_{x'\sim\mathbb{P}(X)}[\delta(P, P')],
\label{eq:tce}
\end{equation}


Based on the causal relation depicted in Figure~\ref{fig:causality_graphs}, we can leverage intervention approaches to interrupt the undesired effects of $Z_c$.  
The intervention is denoted as $do(X: x\rightarrow x')$ and simplified as $do(X)$, where $x$ and $x'$ are inputs before and after an intervention. 
The path $Z_o\leftarrow Z_c\rightarrow A$ forms a backdoor path and thus and thus we aim to perform an intervention to block the path $Z_c\rightarrow Z_o$. 
To achieve this, we consider intervening image $I$ ($\S$\ref{ssec:image_intervention}), text query $Q$ ($\S$\ref{ssec:text_intervention}), or latent embedding $E$ ($\S$\ref{ssec:embedding_intervention}), shown in Figure~\ref{fig:editing} and described in the following subsections. 

\subsection{Image ($I$) Intervention}
\label{ssec:image_intervention}

In Figure~\ref{fig:causality_graphs_image}, the effect of the query text $Q$ is minimized by using a fixed, simple prompt as described in $\S$\ref{ssec:method_causality_analysis}. This allows us to focus on the interactions between $I$, $Z_o$, $Z_c$, and $A$.
We aim to perform intervention $do(I)$ on the image $I$ while assuming that the image after intervention $I'$ minimally changes $Z_c$. Formally, we assume that $P(Z_c \mid I') \approx P(Z_c \mid I)$. This assumption holds if $I'$ primarily affects $Z_o$ and has a much weaker effect on $Z_c$. We expect this to hold in our design, which involves small object modifications and focuses on the region related to $Z_o$.

The confounding effect of $Z_c$ can thus be mitigated using the backdoor adjustment. We can examine the TCE based on Eq.~\eqref{eq:tce} with $P'=\mathbb{P}(A|Q, do(I))$. 
We implement this image intervention $do(I)$ using two different object manipulation designs: pasting a small object in the background of $I$, based on the observation of the differing hallucinatory tendencies of LVLMs regarding the target objects and the rest context elements, as shown in Figure~\ref{fig:editing}; and removing a hallucinatory-inducing object from $I$ where the object is specified by statistics in $\S$\ref{ssec:h_statistics}. 
Specifically, for the {\em image-pasting} intervention, we paste a small image featuring a single object, sized to one-sixth of the shortest side of $I$, at the top left corner. This ensures the object is recognizable and in the background, implicitly affecting  $Z_o$. For the {\em object removal} intervention, we remove one hallucinatory-inducing object in the image based on the highest hallucinatory frequency $\Sigma_{o_i\in O_{h} }P(\,o_i \mid O_n\,)$, where $O_n$ is the non-hallucinatory object set and $O_h$ is its corresponding hallucinatory object set. We report these prior statistics in Table~\ref{tab:Cooccurance_count}. For example, if `person' is a non-hallucinatory object, it is most likely associated with a hallucinatory object; thus, it has the highest priority for removal when present in $I$. We utilize the combination of the GroundingDINO~\cite{liu2023grounding} and IA~\cite{Yu2023InpaintAS} to detect and segment the object and then fill the masked area using the inpainting technique. 


\subsection{Text ($Q$) Intervention}
\label{ssec:text_intervention}

Our proposed text intervention technique comprises two steps, separately prompting for the foreground (FG) and background (BG) generation. This \textbf{foreground-background (FGBG)} technique is designed based on the idea of interrupting the effects from $Z_c$ via changing the object concepts $Z_o$ on the backdoor path, $Z_o\leftarrow Z_c\rightarrow A$ shown in Figure~\ref{fig:causality_graphs_text}. We directly intervene in the text query as $Q'$ with the FGBG strategy and thus $Z_o'$ changes accordingly, paving an alternative causal way to prevent $Z_c$ from affecting $Z_o$.  
The quantified TCE of $Q$ intervention involves the $P(A|do(X))$ probability distribution. 
Our designed \textbf{FGBG} approach is a Chain-of-Thought (CoT)-like prompting technique with the front-door adjustment~\cite{pearl2016causal,zhang2024causal}. By introducing a mediator variable $S$ to perform the two-step prompting, the probability distribution to generate $A$ becomes as below: 
\begin{equation}
    P(A|do(X))=\sum_{S}{P(A|do(S))P(S|do(X))}.
\label{eq:forebackground}
\end{equation}
This approach avoids the intractable access to variable $Z_c$ as the context variable $Z_c$ can hardly be specified. Therefore, we seek an estimation of the term $P(S|do(X))$ and $P(A|do(S))$ for tractable results. Here, $P(S|do(X))$ and $P(A|do(S))$ correspond to the \textbf{FG} and \textbf{BG}, respectively.





First, $P(S|do(X))$ is computed in the path of $X\leftarrow Z_c\rightarrow A\leftarrow S$ between $X$ and $S$. The collision structure of $Z_c\rightarrow A\leftarrow S$ allows us to block the backdoor path and derive $P(S|do(X))=P(S|X)$. 
Second, $P(A|do(S))$ is computed by blocking the path $S\leftarrow X\leftarrow Z_c\leftarrow A$ using backdoor adjustment: $P(A|do(S))=\sum_{X}{P(X)P(A|S, X)}=\mathbb{E}_X[P(A|S, X)]$. 
Instead of navigating the unconstrained space of $X$, we follow the estimation based on the expectation value of $X$~\cite{xu2015show}.
\begin{equation}
    \mathbb{E}_X[P(A|S, X)]\approx P(A|S, \mathbb{E}[X])\approx P(A|S\oplus \mathbb{E}[X]),
\label{eq:p_a_sx}
\end{equation}
where $\oplus$ denotes vector concatenation. We avoid multiple iterations to form a CoT by specifying the prompt to represent $Z_o$ and minimize the variation. Our observed foreground-background description structure and the non-hallucinatory tendency of the first sentence~\cite{zhou2024analyzing} ensure FG to be specific and consistent. We thus empirically use $X$ in a single run to replace $\mathbb{E}[X]$.

Finally, Eq.~\eqref{eq:forebackground} can be rewritten as: $P(A|do(X))=\sum_{S}P(S|X)P(A|s\oplus \mathbb{E}[x])$.
The final intervention result is decomposed into FG $P(S|X)$ and a contextual background (BG) prompt depending on FG, $P(A|s\oplus \mathbb{E}[X])$. Detailed prompts are written in the supplementary.



\vspace{-1mm}
\subsection{Embedding Intervention}
\label{ssec:embedding_intervention}
\vspace{-1mm}
Our embedding intervention approach inspired by model editing research~\cite{cheng2023edit,li2023inference} can generate a direct intervention to $Z_o$ in Figure~\ref{fig:causality_graphs_embedding} without model parameter updates.
Depending on samples $X$ in a proxy dataset with hallucination annotations, we derive a hallucinated group $X_h = \{x \in X| H(f(x)) > 0\}$ and a non-hallucinated group $X_{n} = \{x \in X| H(f(x)) = 0\}$. The corresponding embeddings for $X_h$ and $X_n$ are denoted as $E_h$ and $E_n$. In contrast to other studies aiming to edit specific words, our targeted generative tasks lack fixed ground truth.
Therefore, we propose to measure salient latent embedding dimensions in terms of hallucination and edit these dimensions by retrieving values of the dimensions from non-hallucinated data. 


\noindent
\textbf{Embedding saliency map:}
We measure the saliency dimensions over the whole sequence via statistical significance. Specifically, we utilize the Student’s t-test to examine each dimension between $E_h$ and $E_{n}$. We select the dimensions with a p-value smaller than 0.001 and derive the saliency maps $M\in \mathbb{R}^{T\times D}$ indicating the dimensions statistically significant in discriminating hallucination and non-hallucination groups in the proxy dataset.


\noindent
\textbf{Embedding editing:}
We calculate the distance between a query embedding $E_q$ and the proxy non-hallucinated embeddings $E_{n}$ from $X_{n}$. 
The average embedding $E_k$ of the most similar top-$K$ embedding is then selected through the $l_2$-distance k-nearest neighbor approach. 
$\mathbb{E}_K = \frac{1}{K}\sum_{i=1}^{K}{E_i}$. 
Each query sample obtains the mean embedding from most similar and non-hallucinated samples in the proxy dataset. Then, $E_K$ serves as a non-hallucinated prototype to edit the identified hallucination embedding saliency map $M$. Therefore, the derived embeddings become: $E_q' = (1-\rho)*E_q + \rho*M*E_K$, where $\rho$ denotes a hyperparameter determining editing strength. The embedding $E_q'$ is then used to decode the output texts $A'$. This embedding editing technique is an intervention to $Z_o$ to lessen the effects of $Z_c$ in Figure~\ref{fig:causality_graphs_general}.


\section{Experiments}

We carry out experiments to show the impact of interventions on reducing hallucinations as detailed in $\S$\ref{ssec:image_intervention}, $\S$\ref{ssec:text_intervention}, and $\S$\ref{ssec:embedding_intervention}. Our comparisons are aligned with previous studies on hallucination reduction using the AMBER dataset~\cite{wang2023llm} and COCO validation subset. We test our three intervention approaches on InstructBLIP~\cite{dai2024instructblip} and mPLUG-Owl2~\cite{ye2023mplug} using the default parameter settings of the original paper. For mPLUG-Owl2, we set the hyperparameters temperature and max new tokens to 0.7 and 512, respectively, to derive similar results in AMBER benchmark~\cite{wang2023llm}.

\textbf{Baselines:} 
We compare our approach with the following baselines:
(1) Opera~\cite{huang2023opera}: reduces hallucinations by lowering the attention weight on the summary tokens based on observation occurring on these tokens.
(2) VCD~\cite{leng2023mitigating}: distorts detected hallucinated objects and performs contrastive decoding to avoid generating these objects.



\begin{table*}[t]
\centerline{
\small{
\setlength{\tabcolsep}{0.3mm}
\begin{tabular}{@{}ccccccccccccccc@{}}
\toprule
LVLM         & \multicolumn{7}{c}{InstructBLIP}                                                      & \multicolumn{7}{c}{mPLUG-Owl2}                                            \\
\cmidrule(lr){2-8}\cmidrule(lr){9-15}
Dataset      & \multicolumn{4}{c}{Amber}                     & \multicolumn{3}{c}{COCO}            & \multicolumn{4}{c}{Amber}                     & \multicolumn{3}{c}{COCO} \\
Metrics      & CHAIR$\downarrow$ & Cover$\uparrow$ & HAL$\downarrow$ & Cog$\downarrow$ & CHAIRs$\downarrow$ & CHAIRi$\downarrow$& Recall$\uparrow$ & CHAIR$\downarrow$ & Cover$\uparrow$ & HAL$\downarrow$ & Cog$\downarrow$ & CHAIRs$\downarrow$      & CHAIRi$\downarrow$ & Recall$\uparrow$    \\ 
\cmidrule(lr){1-1}\cmidrule(lr){2-5} \cmidrule(lr){6-8}  \cmidrule(lr){9-12}  \cmidrule(lr){13-15}
Baseline     &   9.0     &   52.4    &  38.8       &  4.5        &  58.6      & 16.1        & 73.1         & 9          &  53         &  40.4      & 5         & 60.2   &   17.5 &   76.1     \\
Opera         &  8.5     &   52.6    &   37.5      &  4.0        &   41.0     & 11.2        & 71.7         &  9         &  49.8       &  36.1      & 3.7       & 48.8   & 15.0 & 71.6  \\
VCD (sampling) &  20.3   &  52.1     &   58.5      & 7.1         &  54.6      & 25.8        & 61.7         &9.9         &  52.6       &  42.4      & 5.3       &63.0    &  18.3 & 76.6 \\
\midrule
Image pasting   &   5.5  &   47.8    &  27.2       &2.7          &  40.4      & 12.3        & 68.5         & 6.4        &   44.7      &   29.0     & 2.5       & 49.6   &  16.3 &  71.7\\
object removal  &  11.7  &   \textbf{56.8}    &  46.2       & 4.1         &   48.8     & 15.0        &\textbf{75.9} &   12.9     &54.0&  46.0      &  4.5      & 59.9   &  19.6  &  \textbf{78.6}     \\
Stopping prompt &   \textbf{5.1}  &  46.7     &  \textbf{20.0}       & \textbf{1.9}         & 21.6       &  8.2        & 61.7         & \textbf{5.5}        &   49.7      &  25.0      & 2.4       & 41.0   &  12.4  &  72.8 \\
FGBG prompt &    5.6     & 53.2          &  27.8       & 2.6& 27.4        &  \textbf{7.0}         & 68.7       &   5.6       &   \textbf{54.3}     & 26.5      &2.1&    30.2     &    \textbf{9.5} &  69.7 \\
BG prmopt      &    6.4  &   52.4    &  27.1      & 2.6         &  \textbf{11.2}      &  8.4       & 41.0          &   7.8      &   39.7      &  \textbf{19.6}       &\textbf{1.7}&   \textbf{20.4}    &14.2   & 43.9 \\ 
Embedding    &    8.1  &   52.7    &  33.6      & 3.8         &  50.7      &  14.9       & 72.4          &   8.6      &   51.9      &  33.4       & 4.2 &   55.6    &  15.7   & 75.4 \\ 

\bottomrule
\end{tabular}%
}
  \vspace{-2mm}
}
  \caption{Hallucination evaluation on AMBER and COCO datasets using InstructBLIP and mPLUG-Owl2. Compared baselines are presented in the upper part and our proposed text and image intervention results are in the lower part.
  }
  \label{table:baseline_intervention}
  \vspace{-2mm}
\end{table*}


\subsection{Evaluation of Image Intervention}
\label{ssec:exp3_image}

We evaluate our \em{image-pasting} and \em{object-removal} intervention methods ($\S$\ref{ssec:image_intervention}) for reducing LVLM hallucinations. The image-pasting intervention uses a single rabbit as the pasted object. 
Table~\ref{table:baseline_intervention} compares our method with two state-of-the-art approaches (Opera and VCD) using two LVLMs (InstructBLIP and mPLUG-Owl2). Our method shows consistent improvements in reducing hallucinations on both LVLMs across the AMBER and COCO datasets and performs comparably to Opera and VCD. This supports our finding that focusing LVLMs away from non-informative backgrounds reduces hallucinations.

We also explore various pasting factors, such as style and semantic relationships, using the COCO dataset; see Table~\ref{table:pasting_exp}.
We find that inpainting, which blends the pasted object into the background, achieves a more consistent style than direct pasting. For semantic pasting and inpainting, we use objects from the same or different supercategories as the original image. Specifically, non-semantic pasting, such as inserting a bird into a kitchen image, performs better by reducing spurious correlations, while semantic inpainting, which inserts objects with similar semantics, tends to increase related hallucinations.


In contrast, the object removal intervention, as shown in Table~\ref{table:baseline_intervention}, results in lower CHAIR and HAL scores but achieves a high Cover rate and a comparable Cog score on the AMBER dataset. Object removal may not effectively reduce hallucinations due to complications like residual background after removing large objects, which can lead to additional hallucinatory artifacts. To better understand the impact of object removal, we also measure its effect on frequently hallucinated objects and their associated inducing words. These findings are detailed in the Supplementary materials.


\begin{table}
\centerline{
  {\small{
    \begin{tabular}{cccccc}
    \toprule
    COCO         &    CHAIRs$\downarrow$ & CHAIRi$\downarrow$& Recall$\uparrow$   \\
    \midrule
    mPLUG-Owl2     & 60.2          & 17.5      & 76.1    \\
    semantic pasting     & 55.4         & 15.8      & 85.2         \\
    Non-semantic pasting &  \textbf{44.8}    & \textbf{12.1}  & \textbf{96.6}   \\
    semantic inpainting    & 63.2 & 19.0    & 78.1    \\
    Non-semantic inpainting     & 51.8 & 14.5    & 89.5    \\
    \bottomrule
  \end{tabular}
  }}
  \vspace{-2mm}
}  
\caption{Results of image pasting intervention with mPLUG-Owl2 on COCO dataset.}
  \label{table:pasting_exp}
  \vspace{-4mm}
\end{table}


\subsection{Evaluation of Text Intervention}
\label{ssec:exp3_text}



We compare the overall hallucination rate and coverage rate in Table~\ref{table:baseline_intervention}. Our proposed \textit{Foreground-Background (FGBG) prompt} can be separated into \textit{FG} and \textit{BG} steps as described in $\S$\ref{ssec:text_intervention}. We additionally consider another baseline:
\textit{Stop prompt} provides a hint for non-existing objects for the model by the prompt: ``There are no [$O_h$] in the image. Then, describe the image.'' The object terms $O_h$ are derived from a prior output of LVLM on the same sample which is compared with the annotated ground-truth. This is a hard upper bound for intervention and suggests the possibility of hardly corrected samples.

In Table~\ref{table:baseline_intervention}, the \textit{FGBG prompt} achieves 5.6 CHAIR, 27.8 HAL, and 2.6 Cog scores using InstructBLIP on the AMBER dataset. Similarly, mPLUG-Owl2 shows low hallucination results with 5.6 CHAIR, 26.5 HAL, and 2.1 Cog scores. The improved results with low hallucination did not compromise coverage, with the highest 53.2 and 54.3 Cover scores using InstructBLIP and mPLUG-Owl2, respectively, indicating a more precise response correction ability compared to other baseline methods. 
Additionally, the \textit{FGBG prompt} outperforms the baselines with significantly lower $CH_s$ and $CH_i$ scores for both LVLMs. InstructBLIP shows a 31.2\% reduction in $CH_s$ and 9.1\% reduction in $CH_i$, while mPLUG-Owl2 demonstrates a 30.0\% reduction in $CH_s$ and 8.0\% reduction in $CH_i$, indicating an overall relative hallucination reduction of around 50\%.

We examine the two-step results of \textit{FGBG prompt}. LVLMs generally respond to the \textit{FG prompt} with short responses similar to image captioning. This leads to notably low CHAIR, HAL, and Cog scores on the AMBER dataset using InstructBLIP (2.2, 4.5, and 0.2) and mPLUG-Owl2 (3.9, 14.1, and 0.7). In contrast, the \textit{BG prompt} induces more hallucinations compared to the FG, with CHAIR scores increasing by 6.4\% and 7.8\%. The higher HAL scores (27.1\% with InstructBLIP and 19.6\% with mPLUG-Owl2) further indicate uncertainty in background descriptions. 
However, \textit{FG prompt} results in a Cover score reduction of 16.5\% and 10\% with InstructBLIP and mPLUG-Owl2, respectively. A similar disadvantage is seen with the \textit{Stopping prompt}, which leads to a Cover score decline of 5.7\% and 3.3\% using InstructBLIP and mPLUG-Owl2, respectively. These results are intriguing, as the LVLMs continue to hallucinate even when informed that certain objects do not exist. 
This suggests a tradeoff between being verbose and conservative for LVLMs. Notably, InstructBLIP tends to replicate foreground descriptions more frequently, resulting in higher Cover scores. However, when combined with foreground descriptions, the results are not superior to those of mPLUG-Owl2.
Another noteworthy observation is that the effects of hallucination reduction are not uniform across each $Z_o$.

\begin{figure*}[t]
\centerline{
    \includegraphics[width=\linewidth]{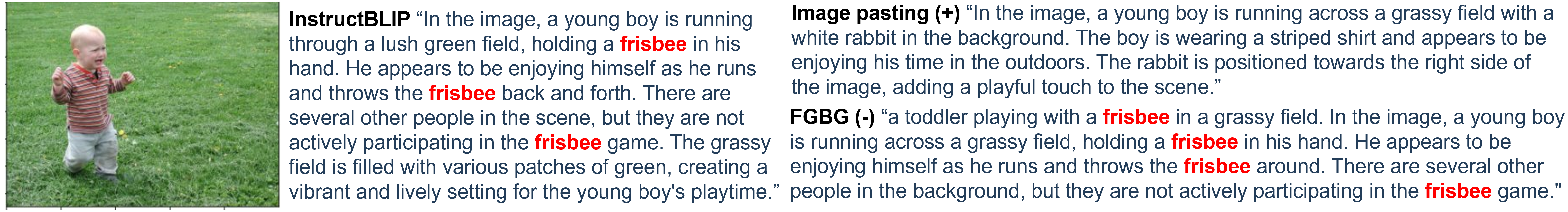} 
  \vspace{-2mm}
}
\caption{A case with the FGBG approach continuing to hallucinate while image-pasting successfully reduces the hallucination, indicating the potential to explore cross-modality casual relations.
}
\label{fig:case_study_frisbee}
\vspace{-0mm}
\end{figure*}

Figure~\ref{fig:case_study_frisbee} illustrates a case where image-pasting eliminates the hallucination, while FGBG continues to hallucinate, suggesting a complex cross-modality causal interaction. To specifically examine the intervention’s effect on the hallucinatory-inducing word, we also measure the change in CHAIR scores for an object $Z_o$ conditioned on the hallucinatory-inducing word. The results are reported in the Supplementary.

\subsection{Evaluation of Embedding Intervention}
\label{ssec:exp3_embedding}

The quantitative evaluation of our embedding intervention approach ($\S$\ref{ssec:embedding_intervention}) is reported in Table~\ref{table:baseline_intervention}. 
Compared to the baseline results on the AMBER dataset, \textit{Embedding} reduces a 5.2\% HAL score while maintaining a 52.7\% Cover score for InstructBLIP. Meanwhile, \textit{Embedding} results for mPLUG-Owl2 sacrifices 1.1\% Cover score but achieves a 7\% reduction in the HAL score. Promising results are observed on the COCO dataset which achieves declined CHAIR scores with kept recall scores.


\begin{figure*}[t]
\centerline{
    \includegraphics[width=\linewidth,height=0.2\linewidth]{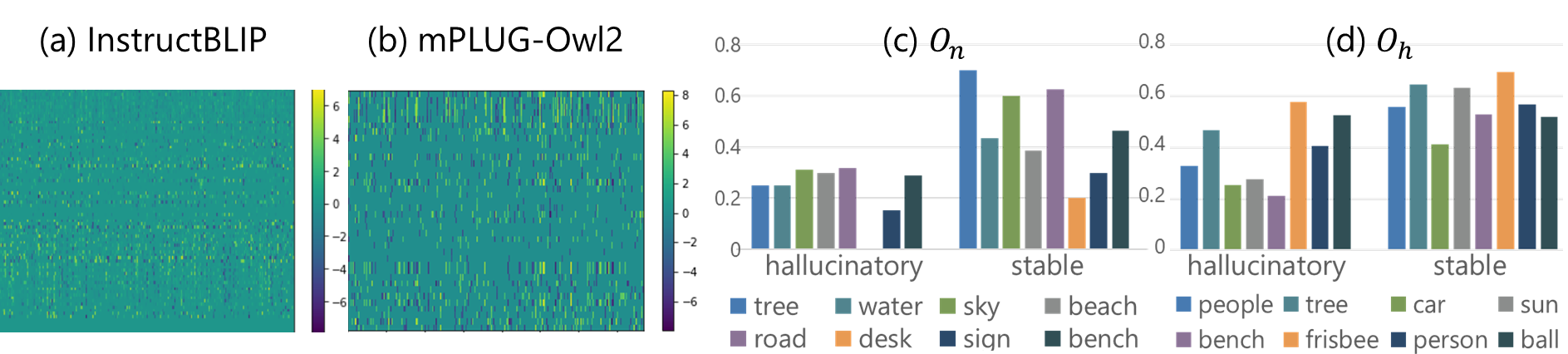} 
\vspace{-2mm}
}
\caption{Embedding Saliency with timestamps in rows and dimensions in columns for (a) Instructblip and (b) mPLUG-Owl2. (c) and (d) show retrieval safe scores given either a non-hallucinatory word ($O_n$) or a commonly hallucinatory word ($O_h$) described in $\S$\ref{ssec:exp3_embedding} using InstructBLIP on AMBER dataset.
}
\label{fig:saliency_map_retrieve}
\vspace{-0mm}
\end{figure*}

  
  



{\bf Saliency visualization:}
%
Figure~\ref{fig:saliency_map_retrieve} visualizes the calculated embedding saliency map $M$ described in $\S$\ref{ssec:embedding_intervention}. 
The salient dimensions can be observed in light green for positive values ($e_h > e_{n}$) and dark blue for negative values ($e_h < e_{n}$) given the index $(i, j)$ with embedding values $e_h=E_h(i, j)$ and $e_{n}=e_{n}(i, j)$. The sparsity of the saliency map illuminates the limited underlying causal feature dimensions affecting hallucination. The visualized results averaging over attention heads are shown in Figure~\ref{fig:saliency_map_retrieve} and the other figures for different heads, timestamps, and dimensions are reported in the supplementary.



{\bf Embedding property analysis:}
%
In the previous experiments, a few data are less likely to be hallucinated which can be used as stable targets to assess the property of an embedding. 
Specifically, we regard a group of data \text{$X_{stable}$} that is never hallucinated with our three text prompts in $\S$\ref{ssec:exp3_text}.
When an arbitrary embedding $E$ retrieves its nearest neighbors $E_0$, we can identify if $E_0$ is in the \text{$X_{stable}$}. 
For a sample set, we obtain a ratio of the retrieved data belonging to \text{$X_{stable}$} and term this ratio as a ``retrieval safe score''. A higher retrieval safe score signifies that this set of embeddings is located near less hallucinated samples.


We regard the cases that comprise either a common hallucinatory-inducing word ($O_n$) or a hallucinatory word ($O_h$) described in $\S$\ref{ssec:h_statistics} narrated in the response. The word can occur in a response that belongs to a group \text{$X_{h}$} being hallucinated with the raw LVLM inference or the stable group \text{$X_{stable}$} less likely to be hallucinated. Figure~\ref{fig:saliency_map_retrieve} (c) demonstrates the results of $O_n$ within the groups of \text{$X_{h}$} or \text{$X_{stable}$} and (d) shows $O_h$ results.

In both Figure~\ref{fig:saliency_map_retrieve} (c) and (d), \text{$X_{h}$} consistently yields lower retrieval safety scores than \text{$X_{stable}$}, indicating distinct properties in the embedding space.
Aside from the general finding, it is necessary to consider the scores under the condition that a given word occurred in that the contexts might be specified differently. Figure~\ref{fig:saliency_map_retrieve} (c) shows that samples mentioned `desk' and `sign' attain low scores in \text{$X_{stable}$} yet still higher than the scores in \text{$X_h$}. A given word of $O_n$ in Figure~\ref{fig:saliency_map_retrieve} (c) exhibits consistently low safe scores for \text{$X_{stable}$} while $O_{h}$ in Figure~\ref{fig:saliency_map_retrieve} (d) includes unexpected high safe scores such as the frisbee. That is, $O_n$ stands closer to hallucinatory embeddings while $O_{h}$ does not. This suggests that non-hallucinatory objects ($O_n$) pose hidden risks for inducing hallucinations, offering new insights into managing hallucinations via embedding space.






\section{Conclusion}

We introduce a causal hallucination probing scheme to analyze potential approaches to mitigate hallucination and identify the underlying hallucination structure that non-hallucinatory objects can induce hallucination.
Our analyses explore image, text, and embedding interventions in the causal framework that can block unwanted causality relations of the inducing objects. 
Our proposed simple approaches achieve significant reduction over other hallucination mitigation methods without model parameter updating. Further, our investigation in embedding intervention uncovers the potential to manipulate the representation space directly. For future works, we will apply causal effects to more complex multimodal tasks and unveil the complex relation in large-scale training and testing datasets. Further exploring the cross-modality causality and model internals in reflecting hallucination under various circumstances is still critical research to extend this study.

{\small
\bibliographystyle{ieee_fullname}
\bibliography{vlm}
}

\end{document}